\pdfoutput=1

\documentclass[11pt]{article}

\usepackage[]{acl}
\usepackage{times}
\usepackage{latexsym}
\usepackage[normalem]{ulem}
\usepackage[T1]{fontenc}

\usepackage[utf8]{inputenc}

\usepackage{microtype}
\usepackage{multirow}

\usepackage{amsmath}
\usepackage{amssymb}

\usepackage{graphicx}

\title{UM6P-CS at SemEval-2022 Task 11: Enhancing Multilingual and Code-Mixed Complex Named Entity Recognition via Pseudo Labels using Multilingual Transformer}

\author{{\bf Abdellah {El Mekki}$^1$} \hspace{0.5cm} {\bf Abdelkader {El Mahdaouy}}$^1$  \hspace{0.5cm}  {\bf Mohammed Akallouch}$^1,^2$\\ \textbf{Ismail Berrada}$^1$ \hspace{0.5cm}  \textbf{Ahmed Khoumsi $^3$}  \\
$^1$School of Computer Sciences, Mohammed VI Polytechnic University, Morocco \\
$^2$Faculty of Sciences Dhar EL Mahraz, Sidi Mohamed Ben Abdellah University, Morocco\\
$^3$Dept. Electrical \& Computer Engineering, University of Sherbrooke, Canada\\
{\tt \{firstname.lastname\}@um6p.ma}
}

\begin{document}
\maketitle
\begin{abstract}

Building real-world complex Named Entity Recognition (NER) systems is a challenging task. This is due to the complexity and ambiguity of named entities that appear in various contexts such as short input sentences, emerging entities, and complex entities. Besides, real-world queries are mostly malformed, as they can be code-mixed or multilingual, among other scenarios. In this paper, we introduce our submitted system to the Multilingual Complex Named Entity Recognition (MultiCoNER) shared task. We approach the complex NER for multilingual and code-mixed queries, by relying on the contextualized representation provided by the multilingual Transformer XLM-RoBERTa. In addition to the CRF-based token classification layer, we incorporate a span classification loss to recognize named entities spans. Furthermore, we use a self-training mechanism to generate weakly-annotated data from a large unlabeled dataset. Our proposed system is ranked 6th and 8th in the multilingual and code-mixed MultiCoNER's tracks respectively.

\end{abstract}

\section{Introduction}
\label{intro}

Recent named entity recognition (NER) models have achieved great performance for many languages and using various benchmark datasets such as CoNLL2003 and OntoNotes 5.0 \cite{devlin-etal-2019-bert}. However, it is unclear whether or not these systems can handle ambiguous and complex entities, especially in the case of short and low-context settings \cite{AUGENSTEIN201761}. It is also unclear weather these systems can be deployed in real-world scenarios where the input data can be in different languages or code-mixed \cite{luken-etal-2018-qed,hanselowski-etal-2018-ukp}. 
In fact, to illustrate these issues, if consider the example of  complex named entities such as the titles of creative works (movies, songs, books ...), they are hard to be recognized by simple NER systems. This is due to their syntactic ambiguity and the form they can take from one context to another. For instance, they can be as an imperative clause  ("Dial M for Murder") or a proposition ("On the beach") which refers to the name of a movie. Thus, it is important to check the performance of NER systems in these scenarios.

The complexity of named entities can be due to three main reasons:
\begin{enumerate}
    \item \textbf{Complex entities:}
    These entities can be represented as complex infinitives (To Kill a Mockingbird) or full clauses  (Mr.Smith Goes to Washington). Additionally, they can be represented as noun phrases or gerunds.  State-of-the-art systems \cite{aguilar-etal-2017-multi} have shown that it is hard to recognize such entities.
    
    \item \textbf{Ambiguous entities and contexts:}
    These types of entities are context-dependent as they can refer to named entities in some contexts, but not in others. “Among Us” which refers to the name of a video game is an example of this challenge. This situation is even more challenging \cite{mayhew-etal-2019-ner} in the case of short sentences with minimal context such as questions or search queries, which most of the time lack some features such as capitalization or punctuation. 
    
    \item \textbf{Emerging entities:}
    This challenge mimics the real-world scenario with many unseen entities, as new named entities are always appearing due to the release of new books, songs, or movies within a short period of time. 
\end{enumerate}

It is well known that the state-of-the-art performance reached by current NER systems is mainly due to the presence of easy entities and well-formed input texts \cite{AUGENSTEIN201761}. Nevertheless, they yield weak performance when applied in multilingual/code-switched input texts having complex or unseen entities. It is also worth mentioning that the reached state-of-the-art results by current NER systems have been achieved using Transformers-based pre-trained language models \cite{devlin-etal-2019-bert} that are known to encode the context of input texts and tokens.

In this paper, we introduce our participating system to the MultiCoNER shared task \cite{multiconer-report}, in particular, the complex MultiCoNER tracks of multilingual and code-switched queries. Our system relies on a deep learning model based on the multilingual Transformer-based pre-trained language model XLM-RoBERTa \cite{conneau-etal-2020-unsupervised}. To handle the complexity of the two tracks, our model is trained using two optimization objectives, as well as for self-training. The main components of our system can be summarized as follow:

\begin{itemize}
    \item \textbf{Optimization of entities span loss:} We train our model to recognize entities spans as an auxiliary task. The aim is to help the model detect the full entities expressions in an input sentence.
    
    \item \textbf{Incorporation of Conditional Random Field (CRF) \cite{10.5555/645530.655813} layer:} We incorporate a CRF layer on top of the Transformer representations to fully exploit the mutual information between tokens in an input sentence.
    
    \item \textbf{ Self-training on unlabeled data:} we create a weakly-supervised data based on the model predictions on unlabeled data.
\end{itemize}

The rest of this paper is organized as follows. Section 2 presents the related work. Section 3 describes the  dataset and the sub-tasks of SemEval-2022 Task 11 \cite{multiconer-report}. In section 4, we present our system overview. Section 5 summarizes and discusses the obtained results for both multilingual and code-mixed tracks. Finally, Section 5 concludes the paper.

\section{Related Work}

During the last years, neural network-based approaches have contributed to improving the performance of NER systems  \cite{panchendrarajan-amaresan-2018-bidirectional,devlin-etal-2019-bert}. This was mainly achieved thanks to word embeddings. Static word embeddings are fed to BiLSTM-CRF models and have helped eliminate manual feature engineering while achieving better performance. On the other hand, Transformer-based Language Models \cite{devlin-etal-2019-bert,conneau-etal-2020-unsupervised} have greatly improved the NER results thanks to their contextualized word representations. 

However, these models may fail when recognizing new or complex entities \cite{luken-etal-2018-qed,hanselowski-etal-2018-ukp}. These challenges are reflecting the real-world setting. Recently, \citet{meng2021gemnet} have proposed an approach to tackle these challenges by using a contextual Gazetteer Representation encoder which can be fused with word-level models. The results have shown that this method enhances the F1-score by +49\% in the uncased setting. This work has been mainly applied to the English language. Finally, in another work, \citet{fetahu2021gazetteer} have explored the code-mixed NER scenario using multilingual Transformers. They have combined the Mixture-of-Experts model with existing multilingual Transformers models to incorporate the multi-lingual gazetteers. The experiments have demonstrated that their proposed approach enhances the F1-score by +31\% in the Code-Mixed NER over the baseline model.

\section{Data}

The dataset of the MultiCoNER (Multilingual Complex Named Entity Recognition) shared task \cite{multiconer-data} is provided to tackle thirteen different tracks, eleven tracks cover the monolingual cases, while the remaining two tracks cover code-mixed and multilingual data. In this paper, we focus on the last two tracks.

The multilingual track covers eleven languages (English, Spanish, Dutch, Russian, Turkish, Korean, Farsi, German, Chinese, Hindi, and Bangla), while the code-mixed track covers a subset of these languages. In both tracks, the entities are annotated into six types: PER, LOC, GRP, CORP, PROD, and CW. Table \ref{tab:train_test_dis} presents the size of the train and test datasets provided for these tracks. We notice that the multilingual track has more data than the code-mixed track, and the test datasets are larger than the train datasets for both tracks. 

\begin{table}[]
\centering
\begin{tabular}{l|l|l}
\hline
\textbf{}      & \multicolumn{1}{c|}{\textbf{Multilingual }} & \multicolumn{1}{c}{\textbf{Code-mixed }} \\ \hline
\textbf{Train} & 168300                                           & 1500                                          \\ \hline
\textbf{Test}  & 471911                                           & 100000                                        \\ \hline
\end{tabular}
\caption{The size of the train and the test datasets for the multilingual and code-mixed tracks.}
\label{tab:train_test_dis}
\end{table}

The provided datasets are labeled using the IOB format which is used for sequence labeling tasks. Table \ref{tab:entities_distro} presents the distribution of the entities and the spans in the train datasets for both tracks.

\begin{table*}[]
\resizebox{\textwidth}{!}{

\begin{tabular}{l|l|lllllllp{2cm}}
\hline
                      &                                  & \multicolumn{8}{c}{\textbf{Classification tasks}}                                                                                                                                                                                                                                              \\ \hline
                      &                                  & \multicolumn{6}{c|}{\textbf{Named entity}}                                                                                                                                                                 & \multicolumn{2}{c}{\textbf{Entity Span}}              \\ \hline
                      & \textbf{Class}                   & \multicolumn{1}{l|}{\textbf{LOC}} & \multicolumn{1}{l|}{\textbf{PER}} & \multicolumn{1}{l|}{\textbf{PROD}} & \multicolumn{1}{l|}{\textbf{GRP}} & \multicolumn{1}{l|}{\textbf{CORP}} & \multicolumn{1}{l|}{\textbf{CW}} & \multicolumn{1}{l|}{\textbf{Entity Span}} & \textbf{Not Entity Span} \\ \hline
\textbf{Mutilingual} & \multicolumn{1}{c|}{\textbf{\%}} & \multicolumn{1}{l|}{22.02}        & \multicolumn{1}{l|}{18.76}        & \multicolumn{1}{l|}{15.0}          & \multicolumn{1}{l|}{14.0}         & \multicolumn{1}{l|}{14.11}         & \multicolumn{1}{l|}{16.12}       & \multicolumn{1}{l|}{38.82}                & 61.17                    \\ \hline
\textbf{Code-Mixed}   & \multicolumn{1}{c|}{\textbf{\%}} & \multicolumn{1}{l|}{18.27}        & \multicolumn{1}{l|}{16.63}        & \multicolumn{1}{l|}{17.88}         & \multicolumn{1}{l|}{13.9}         & \multicolumn{1}{l|}{16.63}         & \multicolumn{1}{l|}{16.69}       & \multicolumn{1}{l|}{41.71}                & 58.28                    \\ \hline
\end{tabular}
}
\caption{The distribution of named entities and entities spans in the train dataset for the multilingual and code-mixed tracks}
\label{tab:entities_distro}
\end{table*}

\section{Methodology}
Before discussing the methodology of our proposed system, we assume that we have $n$ labeled sentences with named-entities $D = \{(x_i^t, y_i^t)\}_{1}^n$ and $m$ unlabeled sentences $U = \{(x_i^u)\}_{1}^m$, where $x_i = \{x_{i,1}, x_{i,2}, ..., x_{i,l} \}$ represents a sentence containing $l$ tokens, while $y_i = \{y_{i,1}, y_{i,2}, ..., y_{i,l} \}$ are the $l$ labels corresponding to these $l$ tokens.

The proposed system incorporates 4 components used on top of the pre-trained Transformer encoder. In the following, we describe each component of our system.

\subsection{Multilingual Transformer encoder}

To encode each word in the input sentence, we use the XLM-RoBERTa (XLM-R) \citep{conneau-etal-2020-unsupervised}. It is a multilingual pre-trained transformer encoder network. We choose to use this encoder for the following reasons: 1) XLM-R is the state-of-the-art encoder in the multilingual and code-mixed settings, and 2) It has been trained on 100 languages including the languages covered in the multilingual and code-mixed tracks. This ensures a good contextualized representation for the input sentences despite their language.

This model was mainly trained using the Mask Language Modeling \cite{devlin-etal-2019-bert}. For an input sentence, 15\% of the words are randomly masked, then the model tries to predict the masked words. As a result of this training process, the model learns the representations of dimension $d$ for the input words of 100 languages that can be fine-tunned on a downstream tasks such as sequence classification or sequence labeling.

\subsection{Span classification module}\label{span_class}

Span classification is a span-wise classifier, where the aim is to classify whether or not, a sequence of tokens are representing a named entity span based on their semantics. It is a binary classification task as the model predicts \textbf{1} if the span is a named entity while it predicts \textbf{0} if not.

Given $H = [h_1, h_2, ...,h_k]$, the vector representations of the $k$ sub-tokens contained in a span $S$,  to learn a representation that encodes all the span tokens, we follow the same approach used in \cite{essefar-etal-2021-cs, el-mekki-etal-2021-bert, el-mahdaouy-etal-2021-deep}, where an attention layer \cite{DBLP:journals/corr/BahdanauCB14} learns the span representations $SH$ based on its tokens, as follows: 

\[ C = tanh(H W^{a}) \]
\[ \alpha = softmax ( C^{T} W^{\alpha})\]
\[SH = \alpha \cdot H^{T} \]

where $W^{a} \in \mathbb{R}^{d \times 1}$, $W^{\alpha} \in \mathbb{R}^{k \times k}$ are the trainable parameters of the attention layer, $C \in \mathbb{R}^{k \times 1}$ and $\alpha \in [0,1]^{k}$ weights the word representations according to their importance for the task at hand.

The vector representations $SH$ of all the spans in an input sentence are then fed to a feed-forward neural network which classifies whether or not, the input representation refers to a named-entity span.

As illustrated in table \ref{tab:entities_distro}, we notice that the majority of spans do not represent a named entity. To tackle this imbalanced data issue, we follow \cite{li-etal-2020-dice} in using the Focal Loss (FL) \citep{8237586}. The FL is given by:

\begin{equation}
    FL(y, \hat{p})
        = - \alpha_y \left(1 - \hat{p}_y\right)^\gamma \log(\hat{p}_y)
\end{equation}

where, $y \in \{0, 1\}$ denotes the span's label, $\hat{p} = (\hat{p}_0, \hat{p}_{1})$ is a vector representing the predicted probability distribution over the labels, $\alpha_y$ is the weight of label $y$, and $\gamma$ controls the contribution of high-confidence predictions in the loss. The performed experiments showed that $\gamma = 0.5$ gives the best result.

As the provided dataset in this shared task has been annotated based on named entities, we adjust it for the span classification task. Therefore, we label all the named entities spans as \textbf{1} while the rest of spans has been labeled as \textbf{0}.

\subsection{NER classification using CRF-Layer}

Most NER systems using Transformers rely on using the first sub-token of each word as input to the classification layer \cite{devlin-etal-2019-bert}. In our system, we follow the work of \cite{acs-etal-2021-subword} in using a pooling of the sub-tokens of each word in the input sentence.

Given $H = [h_1, h_2, ...,h_p]$, the vector representations of the all the sub-tokens contained in a word $w$, an attention layer learns the word pooled representations based on its sub-tokens following the same method explained in section \ref{span_class}.

The pooled representations are then fed to the classification layer which is a Conditional Random Field (CRF) layer in our system. We opt for CRF mainly because the Softmax layer does not take into consideration the dependencies between tokens. The self-attention mechanism, performed by the Transformer encoder, encodes these dependencies in the output vectors of the input sentence. While the CRF which is common in sequence labeling tasks ensures output consistency, it transforms the sequence of input word representations to a sequence of probability distributions, therefore, each label prediction depends on the other predictions in the same input sentence.

\subsection{Self-training}

To take profit from the provided unlabeled dataset in this shared task, we generate a weakly-annotated dataset and re-train the developed model on it. This method has been applied differently in several works \cite{khalifa-etal-2021-self, el-mekki-etal-2021-domain,huang-etal-2021-enhancing}. In our work, we apply the following pipeline:

\begin{enumerate}
    \item We train a model $M$ (based on span classification and NER-CRF explained in the previous subsections) using the provided labeled data $D$.
    \item We use the trained model $M$ to predict the labels of the provided unlabeled data. Then we build a weakly-annotated data $U$
    \item We concatenate the datasets $T$ and $U$ and re-train the sequence labeling model.
\end{enumerate}

It is worth mentioning that during the self-training phase, we remove the span classification module.

\section{Experiments and results}
\label{sec:results}

\subsection{Experimental setup}
We use the PyTorch framework and the Transformers libraries for the implementation of our proposed system. The training of the model is performed on a server with a single Nvidia Tesla P100 with 16GB of RAM. XLM-RoBERTa Large is used as our multilingual Transformer encoder.
Adam optimizer with a learning rate of $1e-5$ is used for all experiments. The system is trained with a batch size of 16 and for 20 epochs.

For the multilingual track, we train our model on the provided labeled multilingual data, then we use the best epoch's model to leverage pseudo-labels from the unlabeled test data, and we re-train the model again from scratch. Besides, for the code-mixed track, we combine the provided data with the training data of the multilingual track and we follow the same training pipeline of the multilingual track.

\subsection{Results}

\begin{table*}[]
\centering
\begin{tabular}{ll|ccc|ccc}
\hline
\textbf{}         &                                & \multicolumn{3}{c|}{\textbf{Multilingual track}}                                                & \multicolumn{3}{c}{\textbf{Code-mixed track}}                                                   \\ \hline
                  &                                & \multicolumn{1}{c|}{\textbf{Precision}} & \multicolumn{1}{c|}{\textbf{Recall}} & \textbf{F1}    & \multicolumn{1}{c|}{\textbf{Precision}} & \multicolumn{1}{c|}{\textbf{Recall}} & \textbf{F1}    \\ \hline
\textbf{Baseline} &                                & \multicolumn{1}{c|}{-}                  & \multicolumn{1}{c|}{-}               & 54.10          & \multicolumn{1}{c|}{-}                  & \multicolumn{1}{c|}{-}               & 58.10          \\ \hline
\textbf{BERT-CRF} &                                & \multicolumn{1}{c|}{68.59}              & \multicolumn{1}{c|}{69.30}           & 68.00          & \multicolumn{1}{c|}{78.47}              & \multicolumn{1}{c|}{78.92}           & 78.78          \\ \hline
                  & \textbf{+ Span Classification} & \multicolumn{1}{c|}{70.71}              & \multicolumn{1}{c|}{70.56}           & 70.25          & \multicolumn{1}{c|}{78.30}              & \multicolumn{1}{c|}{78.77}           & 78.52          \\ \hline
                  & \textbf{+ Self-training}       & \multicolumn{1}{c|}{\textbf{73.21}}     & \multicolumn{1}{c|}{\textbf{72.51}}  & \textbf{72.49} & \multicolumn{1}{c|}{\textbf{79.38}}     & \multicolumn{1}{c|}{\textbf{79.08}}  & \textbf{79.21} \\ \hline
\end{tabular}
\caption{Official Complex NER F1-scores on the multilingual and code-mixed tracks using the proposed system.}
\label{tab:results}
\end{table*}

\begin{table*}[]
\centering
\begin{tabular}{l|lll|lll}
\hline
\textbf{}     & \multicolumn{3}{c|}{\textbf{Multilingual track}}                                                                  & \multicolumn{3}{c}{\textbf{Code-mixed track}}                                                                    \\ \hline
\textbf{}     & \multicolumn{1}{c|}{\textbf{Precision}} & \multicolumn{1}{c|}{\textbf{Recall}} & \multicolumn{1}{c|}{\textbf{F1}} & \multicolumn{1}{c|}{\textbf{Precision}} & \multicolumn{1}{c|}{\textbf{Recall}} & \multicolumn{1}{c}{\textbf{F1}} \\ \hline
\textbf{LOC}  & \multicolumn{1}{l|}{72.24}              & \multicolumn{1}{l|}{81.41}           & 76.01                            & \multicolumn{1}{l|}{80.37}              & \multicolumn{1}{l|}{83.04}           & 81.68                           \\ \hline
\textbf{PER}  & \multicolumn{1}{l|}{85.20}              & \multicolumn{1}{l|}{81.40}           & 83.13                            & \multicolumn{1}{l|}{88.02}              & \multicolumn{1}{l|}{88.67}           & 88.35                           \\ \hline
\textbf{PROD} & \multicolumn{1}{l|}{70.54}              & \multicolumn{1}{l|}{70.24}           & 70.00                            & \multicolumn{1}{l|}{82.68}              & \multicolumn{1}{l|}{80.51}           & 81.58                           \\ \hline
\textbf{GRP}  & \multicolumn{1}{l|}{69.78}              & \multicolumn{1}{l|}{63.6}            & 66.26                            & \multicolumn{1}{l|}{70.85}              & \multicolumn{1}{l|}{73.02}           & 71.92                           \\ \hline
\textbf{CW}   & \multicolumn{1}{l|}{67.58}              & \multicolumn{1}{l|}{68.74}           & 67.95                            & \multicolumn{1}{l|}{75.00}              & \multicolumn{1}{l|}{74.27}           & 74.63                           \\ \hline
\textbf{CORP} & \multicolumn{1}{l|}{73.91}              & \multicolumn{1}{l|}{69.68}           & 71.60                            & \multicolumn{1}{l|}{79.36}              & \multicolumn{1}{l|}{74.99}           & 77.11                           \\ \hline
\end{tabular}
\caption{Official Complex NER F1-scores per entity on the multilingual and code-mixed tracks using the proposed system.}
\label{tab:results_per_label}
\end{table*}

Table \ref{tab:results} presents the submitted results for the multilingual and code-mixed tracks using our system. 
The first row in the table presents the baseline results on the test set published by the shared task organizers. The performance achieved by our best submission largely outperforms the baseline results in both tracks. In fact, the performance using our model boosts the baseline score by 18.39 and 21.11 F1-score points in the multilingual track and the code-mixed track, respectively.
The table also reports the performance of our system during the three stages (ablation study):
\begin{itemize}
    \item The BERT-CRF model that incorporates the named-entity recognition classification layer with CRF, 
    \item The span classification objective,  and 
    \item The self-training. 
\end{itemize}
The results show that the span classification stage enhances the performance of both tracks: the F1-scores achieved using the span classification are 70.25\% for multilingual track  and 78.52 \% for the code-mixed track. However, we notice that the span classification has significantly boosted the F1-score compared to the BERT-CRF model for the multilingual track, while there is a small performance loss  in the case of the code-mixed track.
When performing self-training on the predictions extracted from the model using the span classification stage, a large gain has been achieved in both tracks. For the multilingual track, the F1-score obtained using self-training is 72.49\% with a gain of 3.18\% compared to the system without self-training. For the code-mixed track, our system has achieved the F1-score of 79.21\%  with a gain of 0.87\% compared to the system without self-training.

Finally, Table \ref{tab:results_per_label} presents the performance of our best submission for the multilingual and code-mixed tracks. The proposed system fails the most in predicting the \textbf{GRP} entities for both tracks, meanwhile, it gives its best performance when predicting the \textbf{PER} entities.

\section{Conclusion}

In this paper, we present our Named Entity Recognition (NER) system for complex scenarios on multilingual and code-mixed queries. Our system relies on 4 components: a multilingual transformer encoder, an entity span classification module, a CRF-layer, and a self-training mechanism that leverages information from unlabeled data. We use our system to submit our predictions in the SemEval-2022 Task 11 within the multilingual and code-mixed tracks. The results show that the use of multilingual Transformer and self-training enhances the results in both multilingual and code-mixed cases. Moreover, the incorporation of the span classification module and the CRF layer allow better recognition of complex named entities.


\bibliography{anthology,custom}
\bibliographystyle{acl_natbib}

\end{document}